# A Novel Deep Reinforcement Learning Based Stock Direction Prediction using Knowledge Graph and Community Aware Sentiments

Anıl Berk Altuner[a], Zeynep Hilal Kilimci[b,*]

[a]Department of Information Systems Engineering, University of Kocaeli, Kocaeli 41380.
altuneranilberk@gmail.com

[b]Department of Information Systems Engineering, University of Kocaeli, Kocaeli 41380, Turkey. zeynep.kilimci@kocaeli.edu.tr

*Corresponding author: zeynep.kilimci@kocaeli.edu.tr

**Abstract**

Stock market prediction has been an important topic for investors, researchers, and analysts. Because it is affected by too many factors, stock market prediction is a difficult task to handle. In this study, we propose a novel method that is based on deep reinforcement learning methodologies for the direction prediction of stocks using sentiments of community and knowledge graph. For this purpose, we firstly construct a social knowledge graph of users by analyzing relations between connections. After that, time series analysis of related stock and sentiment analysis is blended with deep reinforcement methodology. Turkish version of Bidirectional Encoder Representations from Transformers (BerTurk) is employed to analyze the sentiments of the users while deep Q-learning methodology is used for the deep reinforcement learning side of the proposed model to construct the deep Q network. In order to demonstrate the effectiveness of the proposed model, Garanti Bank (GARAN), Akbank (AKBNK), Türkiye İş Bankası (ISCTR) stocks in Istanbul Stock Exchange are used as a case study. Experiment results show that the proposed novel model achieves remarkable results for stock market prediction task.

## 1. Introduction

Prediction of stock price index is seen as one of the most challenging applications of time series forecasting. Although there are many studies that address the issues of forecasting the stock price index, most have been associated with advanced financial markets. In particular, there is a limited amount of research in the literature to predict the direction of the stock price index in emerging markets such as Turkey stock exchange. Consistent forecasts made with stock price indices are important for the development of effective market trading strategies [1]. In this way, investors can protect against potential market risks and speculators. In addition, they can have the opportunity to make a profit by trading in the stock index [2]. The stock market is essentially dynamic, linear and non-parametric, complex and chaotic structure, stock market forecasting, financial time series forecasting process is seen as a challenging task [3]. In addition, the stock market is affected by many macroeconomic factors, such as political events, firm policies, general economic conditions, investors' expectations, choices of institutional investors, movements of other exchanges, and investor psychology [4]. Sentiment analysis of financial text has been an important and active research topic for analysts and investors in forecasting stock prices or directions. It has been observed that opinions may affect market dynamics [5].

Knowledge graph is a knowledge base that employs a graph-structured data model in order to consolidate data. Knowledge graphs are often used to store interlinked descriptions of entities such as objects, events, situations or abstract concepts with free-form semantics. The knowledge graph symbolizes a aggregation of interrelated specifications of entities (objects, events or concepts) and their relationships. Knowledge graphs ensure a platform for data consolidation, association, analytics, joining by putting data in context through semantic meta data. Knowledge graphs may employ ontologies as a schema stage. By doing this, they consent logical deduction for revoking implicit knowledge rather than just letting queries demanding explicit knowledge. In this work, we use the knowledge graph when inferring texts/comments with semantic convergence about stocks. Furthermore, social graph is a diagram that illustrates interconnections among people, groups and organizations in a social network. The term is also used to describe an individual's social network [6]. The social graph shape appears as a series of network nodes connected by lines. Nodes on the graph represent an object, and paths between each object are called edges. Edge can be of more than one type, so the link between two objects can be associated. Instead of collecting random users' comments about related stocks, we also use social graphs to gather comments from followers of people with a large number of followers. In this way, we collect comments that contain the thoughts of people who are more relevant or knowledgeable about the related stock.

Sentiment analysis is a classification task that measures the emotional level in the discourse. The opinion can be subjective assessment of something based on personal experience or an aspect for particular issue. Sentiment analysis can be used on finding and extracting the opinionated data on a platform, define subject matter or determine its polarity. Although there are too many classification techniques in order to determine sentiment of opinion, there are very limited state-of-the-art studies on deep reinforcement learning based sentiment analysis. In this study, we propose a novel deep reinforcement learning based stock direction prediction using knowledge graph and community aware sentiments.

Reinforcement learning (RL) is a one of the artificial intelligence techniques that an agent to learn in an interactive environment by trial and error using feedback from its own experiences. There is a situation where the agent is positioned according to the value obtained as a result of every action he makes. The results obtained from the movements can be called reward and punishment. RL offers reward mechanism and create policy for trading. Reinforced learning acquires a behavioral gain by learning this policy through trial-and-error method. Q-learning and SARSA are commonly employed algorithms in many artificial intelligence applications (Alpha Go Zero) and researches [7-8]. Deep reinforcement learning is the combination of reinforcement learning and deep learning that is being able to solve a wide range of complex decision-making tasks that were previously out of reach for a machine to solve real-world problems with human-like intelligence. DeepMind published first successful algorthim about it [9].

In this work, we introduce a novel deep reinforcement learning method to predict the direction of stock prices using knowledge graph and community aware sentiments. With the usage of knowledge graph, semantic convergence about stocks is inferred from comments. Thus, real-world relational objects (human) included to results and using social graphs to get more accurate opinions. It is known that influencers has a strong impact on investment ecosystem. Because of this reason, we gather comments from followers of people with a large number of followers (influencers) with the inclusion of social graph. In this way, we collect comments that contain the thoughts of people who are more relevant or knowledgeable about the related stock instead of collecting unrelated comments of random public users. In this way, we propose a more "live" methodology that includes real world objects and relationships, understands sociological factors and concludes, accordingly. The proposed methodology presents deep reinforcement learning to forecast the direction of stock price by employing knowledge graph-based sentiment signal. For this purpose, deep Q learning technique is used for deep reinforcement learning methodology while Bidirectional Encoder Representations from Transformers (BerTurk) is utilized for sentiment analysis task. DBPedia [10] is also used to construct knowledge graph.

The rest of paper is presented as follows: In Section 2, studies in the literature related to financial applications of deep reinforcement methodologies are explained. Section 3 mentions on the proposed model and its details. Section 4 and 5 advert the experiment results and conclusion part, respectively.

## 2. Related Work

This section provides a brief summary of the state-of-the-art studies on deep reinforcement learning and its financial applications. In [11] Hu and Lin propose deep reinforcement learning model in order to eliminate essential research problems of policy optimization on finance portfolio management. They investigate the impact of recurrent neural network (RNN) models in order to observe the effects of former states and actions on policy optimization. After that, an available risk-oriented reward mechanism is constructed to appraise expected all rewards. Then, authors focus on to integrate reinforcement learning and deep learning approaches in order to find an optimal policy. They report that each type reinforcement learning blended with deep learning method is capable to resolve policy optimization problem. In [12], Rundo introduces deep reinforcement learning approach in order to forecast financial trend in foreign exchange (FOREX) trading system using high frequency trading (HFT) algorithm. Thus, the author proposes the use of an algorithm based both upon long short-term memory network as a deep learning algorithm and on a reinforcement learning methodology for predicting the short-term trend in the currency FOREX market for the purpose of maximizing the return on investment in an HFT algorithm. Authors concludes the study that the introduced method is able to forecast the medium-short term trend of a currency cross based upon the trend of this historically with mean accuracy of nearly 85%.

In [13], Ye et al. present a reinforcement learning based portfolio management system by addressing two main challenges in portfolio management, namely data heterogeneity and environment uncertainty with their proposed model: State Augmented Reinforcement Learning (SARL) framework. The proposed SARL framework boosts the asset information with their price movement forecast as supplementary states in order to combine heterogeneous data and amplify durability against environment uncertainty. To prove the effectiveness of the SARS model, experiment are carried out on two real-world datasets, namely Bitcoin market dataset and High Tech stock market with 7-year Reuters news articles. Experiment result demonstrate that proposed state augmentation approach with SARL provides new foresights and boosts considerably success over standard RL-based portfolio management methods and other traditional techniques. In [14], Xiao and Chen propose sentiment analysis-based reinforcement learning model to predict the stock return using only text data. Q-learning technique is performed as a reinforcement learning methodology in order to discover the optimal trading policy by learning the feedbacks from the market. Experiment results show that both of the machine learning method and the Q-learning technique excels the traditional model, logistic regression, without sentiment attributes. They conclude the paper that forecasting direction of stock price using only text data from Twitter sentiment is challenging but encouraging.

In [15], Li et al. propose a method based on a deep reinforcement learning methodology for trading stock and forecasting direction of stock price. Three different and conventional deep reinforcement learning techniques are utilized, namely deep Q-network (DQN), double deep Q-network (DDQN), dueling double deep Q-network (DDDQN). To show the effectiveness of deep reinforcement learning methodology, the historical daily prices and volumes of random ten stocks in all US-based stocks and exchange trade funds (ETFs) are chosen from NYSE, NASDAQ, NYSE MKT to construct the dataset. Experiment results indicate that the usage of DQN for direction of stock price outperforms other deep reinforcement learning techniques. In [16], Koratamaddi et al. present a new deep reinforcement learning methodology in order to construct an automated system for trading purpose. For this aim, five different models are employed namely, min variance analysis, mean variance analysis, deep deterministic policy gradients, adaptive deep deterministic policy gradients, adaptive sentiment-aware deep deterministic policy gradients. In addition to including historical stock price data, authors also investigate the impact of market sentiment comprising of the Dow Jones companies between 2015 February and 2018 February. Authors report that adaptive sentiment-aware deep deterministic policy gradients technique outperforms other models to orient investments. In [17], Chen and Gao introduce a deep reinforcement learning technique for automated stock trading. Deep Q-network (DQN) and Deep Recurrent Q-network (DRQN) are evaluated to automate stock trading system. In order to observe the efficacy of the proposed model, the S&P 500 ETF and its daily movements are employed as a dataset. Authors state that DQN model performs better than DRQN technique to trade the stocks. In [18], Nan et al. introduce reinforcement learning model by blending with sentiments of news headline and knowledge graph for trading purpose. Deep Q-network is assessed as deep reinforcement methodology. To prove the effectiveness of the DQN network, stock data of Microsoft, Amazon, and Tesla and text data from the Reuters account on Twitter are gathered from January 2018 to December 2018. They conclude the paper that the reinforcement learning methodology outperforms other models in terms of profits.

## 3. Methodology

In this section, a brief summary of the methods, materials, and proposed framework are presented.

### 3.1. *Bidirectional Encoder Representations from Transformers (BERT)*

Bidirectional Encoder Representations from Transformers (BERT) is a new generation word embedding model, which means bidirectional encoder representations. Unlike other word embedding models of the BERT model, it is designed to pre-train the dataset in both layers in two directions and to condition the word in both right and left contexts [19]. The BERT model can be used to fine-tune with an additional output layer to create cutting-edge models without the important task of answering questions and language extraction. It is conceptually simple and empirically powerful [19]. In Figure 1, the architecture of BERT model is presented where the arrows indicate the information flow from one layer to the next. The T1, T2, T3 boxes at the top indicate the final contextualized representation of each input word. Input words are demonstrated with E that lies from 1 to n.

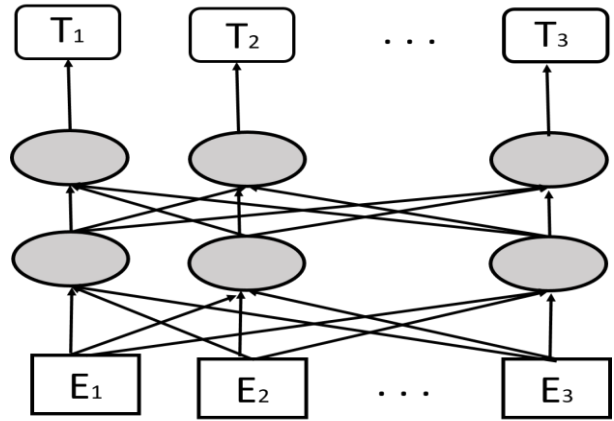

**Fig 1.** The architecture of BERT model.

### 3.2. *Deep Q Network (DQN)*

The advantages of neural network and Q-learning algorithm are blended for DQN [20] approach. It is obtained by enhancing the function of experience replay from the transition of former state (experience) in the random sample training and accomplishing correlated and unsteady distribution data. In DQN, the present learned experience is demonstrated with the Q-value. The logic behind of DQN method is to learn the function of q-value and precisely forecast the q-value of each action in different states. The Q value is actually a score acquired by the agent owing to coaction with the environment and its self-experience, called as the target Q value. In summary: As a first step, samples are collected and stored in a replay buffer with present policy. Second, random sample batches of experiences are provided from the replay buffer, called as experience replay. Random sample experiences are employed in order to eliminate highly correlated experiences of past sequential experiences at this step. Thus, major bias issues are eliminated that can appear from correlated data. Third, the sampled experiences are employed in order to update the Q network. For this purpose, the mean square error (MSE) between the target value of Q and our current output Q is minimized according to equation of Bellman in order to update the Q network. Then, all steps (1-3) are repeated until the target q-value is reached.

### 3.3. *Double Deep Q Network (DDQN)*

In deep Q and Q learning methodology, optimal Q value is employed in order to pick up and evaluate an action. For this reason, selecting an overestimated value can induce to overestimation of the real value of Q. Assessing the maximum overestimated values is implicitly evaluating the forecast of the maximum value. Hereby, this over forecast presents a maximization bias in learning methodology. This over forecast can be problematical because Q-learning contains bootstrapping, namely learning forecasts from forecasts. Thus, conventional DQN tends to notably over forecast action values by causing unsteady training procedure and low-quality policy. In order to avoid the overestimation problem of DQN, double deep Q network approach is proposed by [21] by utilizing two individual Q-value forecaster, each of which is employed to update the another. With the utilization of these separate forecasters, Q-value forecasts has the ability to predict the actions picked up employing the opposite forecaster. Thus, the problem of maximization bias is prevented by delivering updates from biased forecasts.

### 3.4. *Dueling Double Deep Q Network (DDDQN)*

DDDQN [22] is constructed by aggregating double DQN and dueling DQN approaches. While DDQN is mentioned before the details about dueling DQN is given at first. In dueling DQN, there are two different estimates. The first one is to forecast for the value of a given state while second one is to predict for the advantage of each action in a state. Thus, q-value is denoted as how advantageous it is to be at that state and putting into an action at that state. Thence, q-value of an action at that state is dissociated as the total of the value of being at that state and the advance of putting into that action at that state. With decomposing procedure of the forecast, dueling DQN is capable to learn which states are worthy or not without having to learn the impact of each action at each state. In DDDQN, the mean advantage of all actions feasible of the state is subtracted from the output of aggregation layer. Thus, the problem of identifiability is eliminated that causes problem for back propagation that means not to find two elements. One forecasts the benefit for each action and the other forecasts the state value. This means that advantage function estimator is enforced have no advantage at the selected action.

### 3.5. Proposed Framework

In this work, we propose to feed the reinforcement learning methodology with deep learning and sentiment analysis in order to predict the direction of stock prices. The relations of the stock/company and opinions about the stock/company are of top priority for the proposed model. At least as important as, this is the relevance of the people who give their opinions to the investment. For this purpose, community finding algorithms are developed. Thus, investor community and influencer scores are constructed on the social graph. Moreover, an information/knowledge graph is also employed on the expansion of the assets that are going to draw an opinion on. For this purpose, four basic types are employed such as locationCountry (country where the company is located), keyPerson (people important to the company), product (company's products), parentCompany (a parent company of the company), subsidiary (companies owned by the company). After gathering all keywords, tweets are collected through influencer listener services. These tweets include polarity extracted from the sentiment analysis model. The sentiment analysis model is developed utilizing the fine-tuned BERT model. Then, biased value is acquired by multiplying this polarity by the normalized value which is given to reinforced learning as a signal. Finally, a novel deep reinforcement learning based stock direction prediction model is constructed using knowledge graph and community aware sentiments.

#### 3.5.1. Social graph and dataset construction

In order to construct a social graph, Twitter data is gathered. To create a scattered distribution, the person who will start the scrape must be a high-profile account that appeals to the general audience. It is very important that the social graph obtained for determining various target groups has a scattered distribution. After the scrapping process is completed, a graph structure that expresses the general mass is obtained. There are many solutions for finding a community after constructing a social graph. Our solution will be to create a graph with the followers of the best influencers in the selected investment area and score it according to the number of followers in that graph. For this purpose, node score is calculated with follower count, named as influencer score. This approach is carried out on the community and extracted the influencer score of them. Because it is necessary to set a constraint to determine influencers with a large number of followers, having more than 100 followers within the community is the threshold for obtaining the top influencers in this study. As a result, 580 influencers are taken into consideration in this work. Scrape process of the tweets is performed between January 2015 and December 2020 utilizing Twint, which is an advanced Twitter scraping tool written in Python that allows for scraping Tweets from Twitter profiles without using Twitter's API. Similar to Twitter dataset, the time series data is gathered between January 2015 and December 2020 using Yahoo Finance API.

#### 3.5.2. Knowledge graph

Ontologies are databases of terms connected with semantic relationships. They are often represented in a graph with entities and relationships. Knowledge graph, are more complex graphs where the entities are connected with features of their own. With these features, entities establish semantic connections between each other in a way to define each other on the basis of features. In this work, we use DBPedia knowledge base that describes 4.58 million entities [10]. Linked real word entities can establish relational affinity and find out which entity is relevant to which entity. In traditional methods, while this type of tweet analysis is associated with the keyword search, we will search for the keywords together with the relational keywords rather than just searching directly, and we will analyze not only the mentioned keywords but also the tweets that may be related.

Considering one of datasets, Garanti Bank, in order to reveal close entities, we first determine the relationship types manually and add other entities to the keyword dictionary as a result of these types. The relationships consist of 5 basic types, namely location types, person types, parent companies, subsidiaries, and products. DBPedia relationships are presented for each type category as below:

- Location Types → LocationCountry, RegionServed

- Person Types → KeyPerson, KeyPeople

- Parent Companies → parentCompany

- Subsidiaries → subsidiary

- Products → product

### 3.5.3. Sentiment analysis

Sentiment analysis is a method of determining whether a piece of writing, text, document is positive, negative or neutral. It is also known as idea mining, which examines the idea or attitude of a speaker. The common use of this technology is to discover how people feel about a particular subject. It refers to the use of intellectual mining, natural language processing, text analysis, computational linguistics and biometrics to systematically identify, measure and analyze emotional states and subjective information. Generally expressed emotion analysis aims to determine the attitude of a speaker, author or other subject regarding a topic, general contextual polarity or emotional response to a document, interaction or event. This attitude can be a trial or evaluation, or intended emotional communication. Our sentiment model is based on fine-tuned Turkish BERT model . Fine-tuned model is trained with e-commerce dataset which size 50k.

### 3.5.4. Sentiment support score calculation

Before the sentiment analysis, we need to obtain the effect coefficient for each tweet. The effect coefficient is a type of coefficient associated with the keyword with each tweet, which influencer is tweeted, favorited count, mention count it is, retweet count. Interaction types can be different each other. Some types support values have more than others. "Retweet" is mentioned that by sharing the same thought, it kind of supports that idea. For calculating efficient score (ES), we use all type of interaction and influencer score (IS). Interaction bias (IB) is acquired by summing all biases from every interaction type. Retweet bias (RB) has stronger than other types, because of that we use biased retweet score instead of plain retweet score (RC). ES calculation process can be different based on searched keyword type. In previous section, we talk about knowledge graph for relational entity to our main entity. If searched tweet's score, ES is divided by 4, its value will be reduced. Finally, for the main entity, Equation 4 is used for sentiment support with main entity.

$$RB = RCoe + RC \tag{1}$$

$$IB = RB + LC + RepC \tag{2}$$

$$ES_{rel} = (IB + IS) / RP \tag{3}$$

$$ES = IB + IS \tag{4}$$

where RCoe refers to retweet coefficient and RC is retweet count, LC denotes like count, RepC presents reply count, and RP is reduced parameter. After extracting efficient coefficient, each tweet is processed in the sentiment model. The positive, negative, and neutral values are multiplied by 1, -1, and 0, respectively. Since the effect coefficient range increases from very low to very high numbers, vector norm of array is found and normalized. When the stock prices are considered, a normalization is carried out between 0 and 100. The results are collected cumulatively for each day and recorded in the database. In this way, it is determined how the sentiment is analyzed for that day.

### 3.5.5. Markov Decision Process

In proposed deep reinforcement learning methodology, stock direction prediction system is constructed by associating time series analysis with sentiment analysis results. Deep reinforcement learning can be summarized as creating an algorithm or an artificial intelligence (AI) agent that learns to interact directly with an environment by utilizing reward/penalty mechanism. In this way, the AI agent like a person learns the results of their actions (reward or penalty) rather than being taught explicitly. In the proposed system, the reward mechanism is a numerical value resulting from the sentiment analysis and the signals of some indicators, apart from the traditional time series analysis. It is observed that direction prediction of stocks is a sequential decision-making process, as the investor would require to make investment options every day, one day after the other. Thus, the issue of direction prediction of stocks is modelled as a Markov Decision Process (MDP). Our current environment daily defines each state utilizing six variables:

    i.    Closing stock price on today's date.

    ii.    Sentiment value towards the stock for today's date.

    iii.    Average growth value of last 5 days.

    iv.    Average sentiment value of last 5 days.

    v.    Average growth value of last 30 days.

    vi.    Average sentiment value of last 30 days.

While (i) value is essential for maintaining the state of the agent (ii), (iv), and (vi) ensure the sentiment information calculated as described in the previous section over a short time period (5 days) and a longer time period (30 days). In the reward formulation, there are three basic time intervals that are zero to five days, five to thirty days, and time interval greater than 30 days. The effect coefficient of daily values is set to 2 times higher than the last 5 days and 4 times higher than the last 30 days because the current issues have a greater effect on the stock market. Growth information is provided with the state of the agent (iii), and (v) by calculating growth bias. It is effective in equal coefficients since there is no up-to-date growth rate on the stock. For (iii), the average of the last 5 days is taken and subtracted from the closing value of that day. The difference is divided by the average of 5 days and growth is obtained according to the average. To find the effect of this rate on that day, a signal is obtained by multiplying by that day. After that, the reward is sent to the environment as a signal by blending time series data, sentiment analysis, and time series indicator in Markov decision process as mentioned before in order to get an action. In this way, a policy will be determined based on the time series analysis combined with both growth and sentiment bias.

In action space, the agent, the proposed stock prediction model, interacts with this environment on a per day basis. It has the option to take three actions:

- Buy a stock (BS)
- Sell a stock (SS)

- Do nothing/Hold. (DNH)

In order to construct Deep Q-Network (DQN) architecture, Q-network, and target network is employed with each with 3-hidden layers for the purpose of function approximation. Each hidden layer is constructed with 64 units and ReLU activation function. The input layer had six input nodes that corresponds to feature of each state. Finally, DQN architecture is completed with three nodes that correspond to the actions namely, BS, SS, and, DNH. In addition, the experience replay buffer size is set to 1000. Training procedure with DQN is performed by mini-batch gradient descent using Adam optimizer [23]. The agent is interacted with the environment between January 1, 2015 and December 31, 2020. The agent in the form of a DQN is trained over 50 epochs. The training procedure is carried out on Garanti Bankası, Türkiye İş Bankası, Akbank datasets. In Figure 2, a general flowchart of the proposed framework is given.

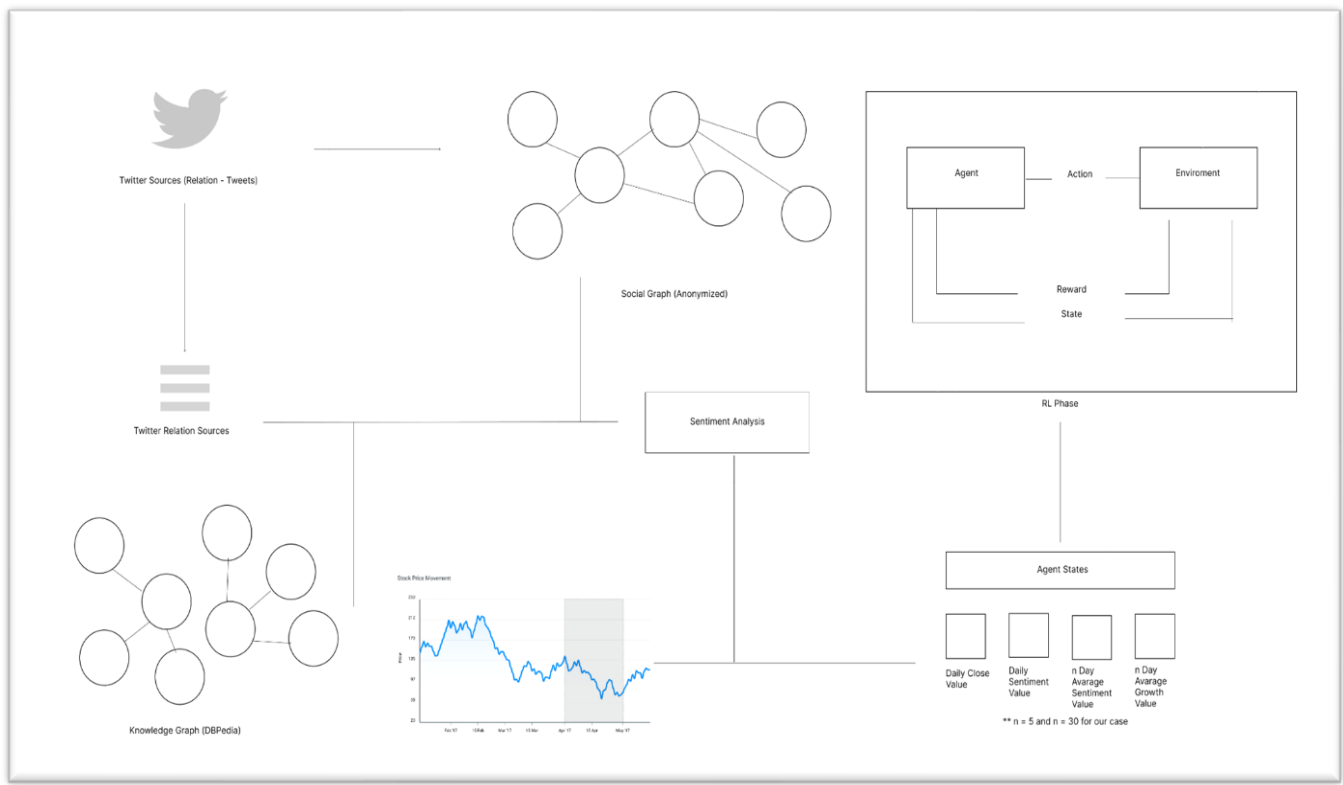

**Fig 2.** A general flowchart of the proposed framework

## 4. Experiment Results

The primary purpose of this work is that maintaining sentiment information to the agent on a daily basis would contribute to its performance and it would be able to ensure more profit by determining buy or sell signal. For this purpose, we compare deep Q network with and without community analysis, double deep Q network with and without community analysis, dueling double deep Q network with and without community analysis approaches. In other words, it is proposed to compare the performances of an agent with sentiment data provided and another agent without any sentiment data provided. The following abbreviations are utilized for methods used in the experiments: DQN: deep Q network, CA-DQN: Deep Q network with community analysis, DDQN: Double deep Q network, CA-DDQN: Double deep Q network with community analysis, DDDQN: Dueling double deep Q network, CA- DDDQN: Dueling double deep Q network with community analysis, GARAN: Garanti Bankası dataset, ISCTR: Türkiye İş Bankası dataset, AKBNK: Akbank dataset. All techniques are implemented on Google Colab environment provided free GPU utilization by Google. The best profit results are represented in bold letters in the table. For Garanti Bank, the dictionary set keywords are "Turkey", "Ferit Şahenk (Founder)", "Doğuş Holding (Parent Company)", "BBVA (Parent Company)", "#garan (Exchange Code)", "Garanti Bank (Company Name)". For Akbank, the dictionary set keywords are composed of "Turkey", "Suzan Sabancı Dinçer (Founder)", "Personal Bank Services (Service Area)", "Investment Banking (Service Area)", "Mortgage (Service Area)", "#akbnk (Exchange Code)", "Akbank (Company Name)". For Isbank, the dictionary set keywords are constructed as "Turkey", "Investment Banking (Service Area)", "Special Banking (Service Area)", "#isctr (Exchange Code)", "İş Bankası (Company Name)". In Table 1, performance of different agents for Garanti Bankası, Türkiye İş Bankası, Akbank datasets are demonstrated in terms of profit score. It is clearly observed that the proposed framework by including community analysis exhibits higher performance compared to the traditional techniques such as DQN, DDQN, and DDDQN. CA-DQN outperforms both other versions of CA-based techniques and traditional reinforcement learning methodologies for AKBNK and ISCTR while the best profit score is obtained with the utilization CA-DDDQN for GARAN dataset. The profit score order can be summarized for GARAN dataset as CA-DDDQN> CA-DDQN> CA-DQN> DDDQN> DQN> DDQN. In general, even performance of profit order varies among datasets, the proposed CA-based proposed techniques perform well compared to the conventional models. As a result of Table 1, the results demonstrate that all CA-based proposed techniques can effectively learn a profitable strategy from history data.

**Table 1**
Performance of different agents for different stocks in terms of profit scores.

| DATASET | METHODS | | | | | |
|---------|-----|--------|------|---------|-------|---------|
|         | DQN | CA-DQN | DDQN | CA-DDQN | DDDQN | CA- DDDQN |
| GARAN   | 7   | 9      | 1    | 376     | 23    | **1286** |
| AKBNK   | -5  | **138** | 3   | 16      | 10    | 5       |
| ISCTR   | -17 | **184** | 4   | 150     | 3     | 20      |

Figure 3, Figure 4, and Figure 5 show the reward and loss functions of the GARAN, AKBNK, and ISCTR stocks, respectively after the training by six models. Aforementioned before, reward specifies the target in the reinforcement learning problem. Reward function denominates to the overall reward caused by the modification of environmental state affected by the sequence of actions chosen at each time period. In other words, it is an instant reward that can evaluate the cons and pros of the actions. In addition, loss function is employed to forecast the grade of discrepancy between the actual value and the forecasted value of the model. The experiment compares the loss functions of the six deep reinforcement learning models in order to evaluate and compare the economic advantages of these techniques. In Figure 3, Figure 4, and Figure 5, (a), (b), (c), (d), (e), and (f) notations specify DQN, DDQN, DDDQN, CA-DQN, CA-DDQN, and CA-DDDQN models, respectively.

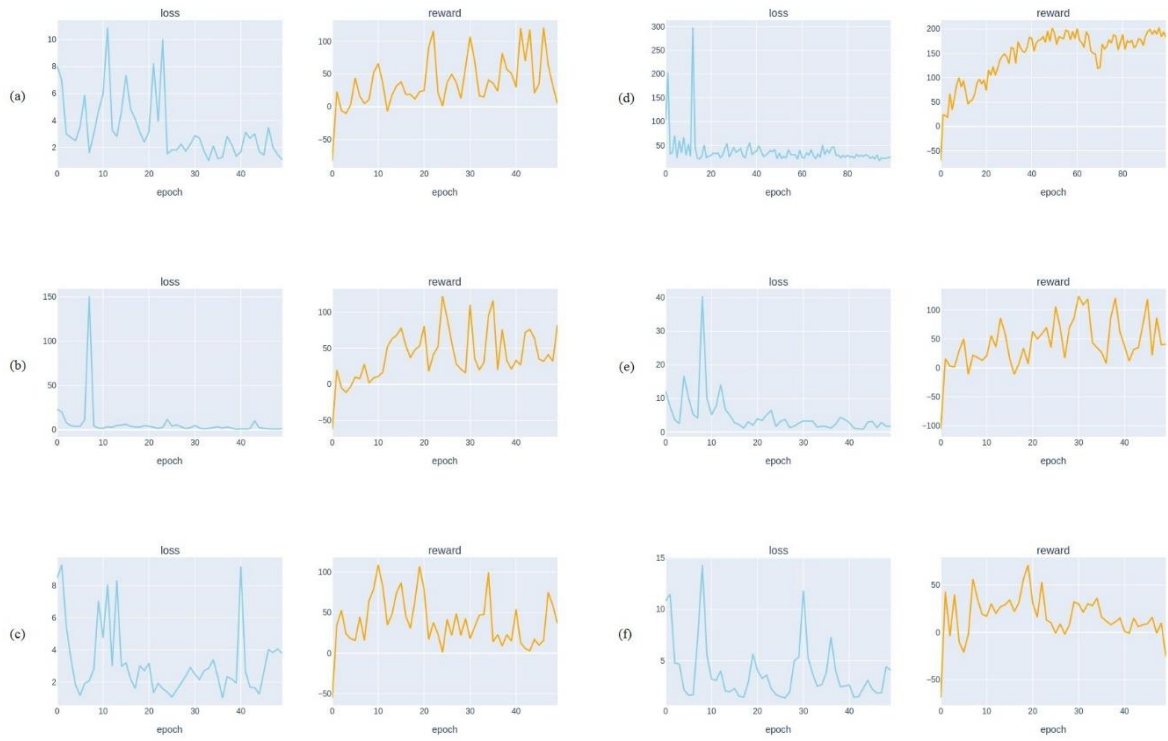

**Fig 3.** The loss and reward function of the GARAN stock.

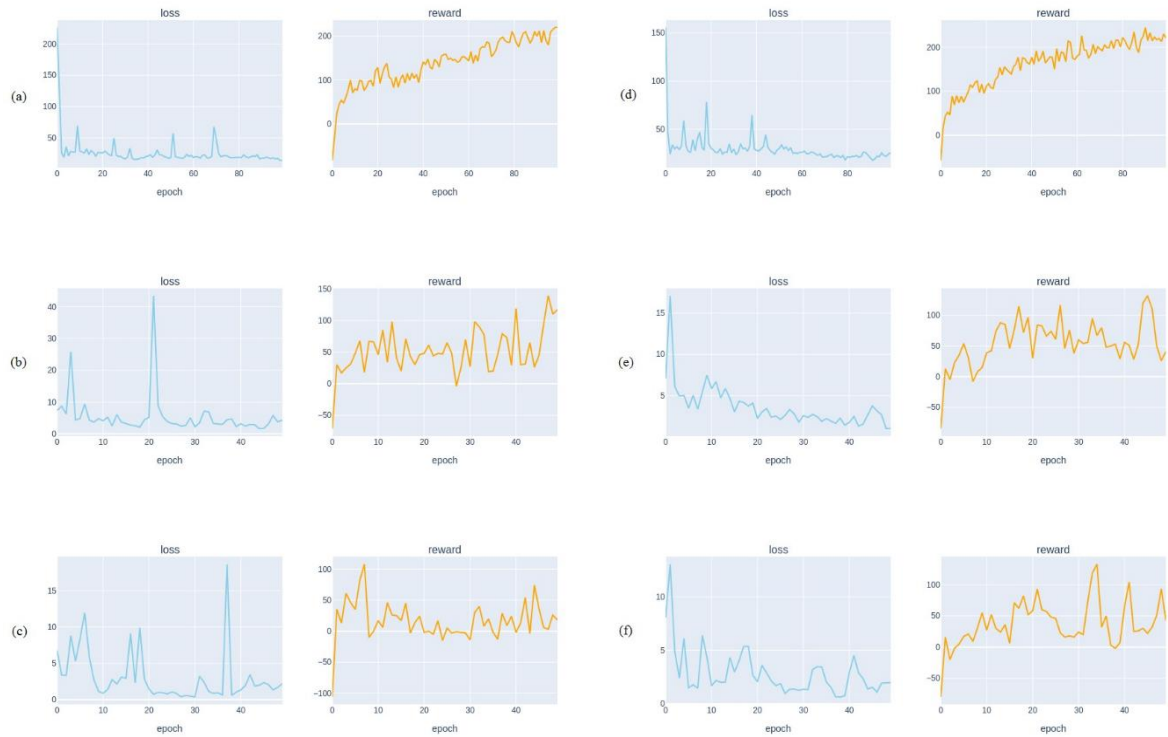

**Fig 4.** The loss and reward function of the AKBNK stock.

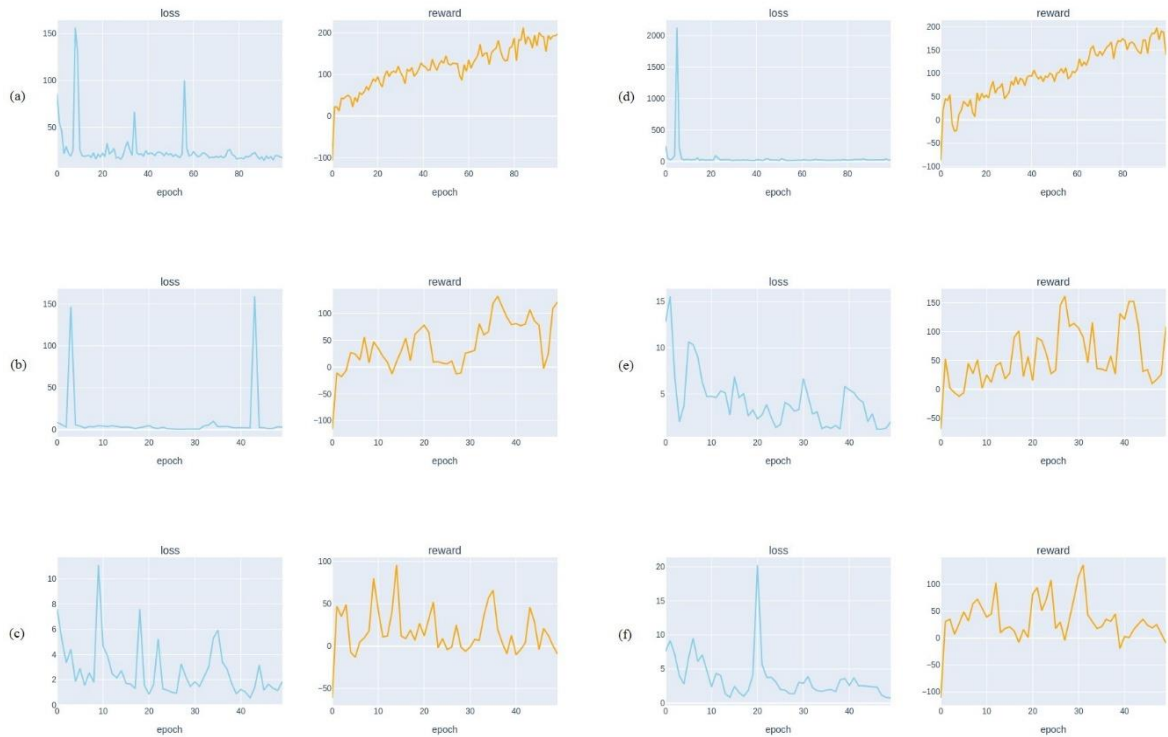

**Fig 5.** The loss and reward function of the ISCTR stock.

In Figure 6, performance of the best agent for GARAN, AKBNK, and ISCTR stocks are presented. Figure 6 shows the close price forecasting line chart of the above three stocks, where the black line indicates the original close price, and the red line demonstrates the predicted close price. In every subfigure, x-coordinate represent the date, while the y-coordinate represents the corresponding closing price of each stock. For all datasets, we evaluate test dataset, which spans from January 1, 2020 to December 31, 2020 to demonstrate the contribution of CA-based novel deep reinforcement learning approaches for predicting direction of stocks in Istanbul Stock Exchange. CA-DDDQN performs very well in GARAN dataset while CA-DQN approach is almost overlapped to the closing price, which means CA-based techniques are efficiently capable to learn a profitable strategy from history data. This means that including the knowledge graph and community aware sentiments remarkably contributes the direction of prediction stocks to orient the investments for investors, analysts, researchers. Furthermore, blending knowledge graph and community aware sentiments with deep reinforcement learning methodology becomes the stock price estimations more stable and robust compared to the traditional reinforcement learning approaches. Considering the prediction performance of and flexibility of the proposed framework, it can be also applicable other investment tools such as digital currencies, stocks, mineral commodities such as gold, silver, bonds, funds and such products. That is why the model we propose for investors, researchers and analysts becomes even more attractive.

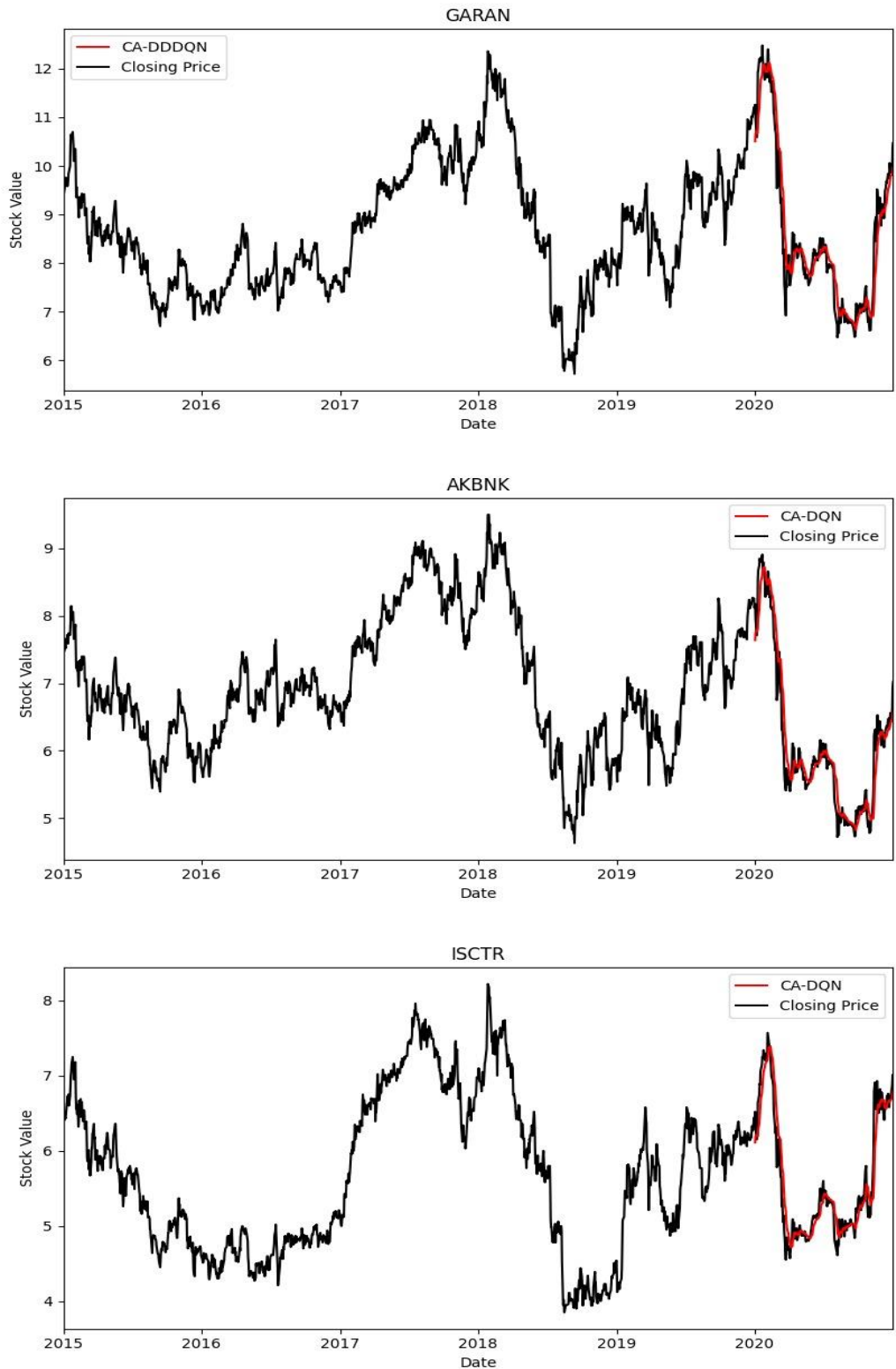

**Fig 6.** Comparison of the best proposed approaches and closing price on the GARAN, AKBNK, and ISCTR stocks

## 5. Discussion and Conclusion

In this work, we propose a novel technique that is based on deep reinforcement learning methodologies for the direction prediction of stocks blending sentiments of community and knowledge graph. For this purpose, a social knowledge graph of users is firstly constructed thereby evaluating relations between connections. Then, time series analysis of related stocks and sentiment analysis is combined with deep reinforcement methodology. Turkish version of Bidirectional Encoder Representations from Transformers (BerTurk) is utilized to analyze the sentiments of the users while deep Q-learning methodology is used for the deep reinforcement learning side of the proposed model to construct the deep Q network. Deep Q network, proposed deep Q network with community analysis, double deep Q network, proposed double deep Q network with community analysis, dueling double deep Q network, proposed dueling double deep Q network with community analysis approaches are employed in the experiments to compare the prediction performance of each model. In order to demonstrate the effectiveness of the proposed model, Garanti Bank (GARAN), Akbank (AKBNK), Türkiye İş Bankası (ISCTR) stocks in Istanbul Stock Exchange are used as datasets. Experiment results demonstrate that the proposed novel models perform remarkable forecasting results for stock market prediction task. As a future work, we also intend to blend transfer learning strategy for predicting stock prices by extending investment tools.